\documentclass[sigconf,10pt]{acmart}
\usepackage{graphicx} % Required for inserting images

\usepackage{color}
\usepackage[tight]{subfigure}
\usepackage{comment}
\usepackage{amsmath}
\usepackage{enumerate}
\usepackage{multirow}
\usepackage{hhline}

\usepackage{enumerate}
\usepackage{enumitem}
\usepackage{pifont}

\usepackage{amsmath}
\usepackage{xspace}
\usepackage{algorithm}
\usepackage[font=small]{caption}
\usepackage[noend]{algpseudocode}

\makeatletter
\def\BState{\State\hskip-\ALG@thistlm}
\makeatother

\algdef{SE}[SUBALG]{Indent}{EndIndent}{}{\algorithmicend\ }%
\algtext*{Indent}
\algtext*{EndIndent}
\algrenewcommand\algorithmicindent{1.0em}%

\newcommand{\ie}{\textit{i.e.}}
\newcommand{\eg}{\textit{e.g.}}
\newcommand\subp[1]{\noindent\textbf{#1}}

\usepackage[normalem]{ulem}

\newif\ifcomm
\commtrue %Comment to turn off comments
\ifcomm
\else
\commfalse
\fi
\ifcomm
\newcommand\Wenchen[1]{\textcolor{blue}{Wenchen: #1}}
\newcommand\ran[1]{\textcolor{red}{Ran: #1}}
\newcommand\sv[1]{\textcolor{red}{Shay: #1}}
\newcommand\MM[1]{\textcolor{red}{MM: #1}}

\newcommand\revold[1]{\textcolor{red}{#1}}
\else

\newcommand\revold[1]{}
\newcommand\Wenchen[1]{}
\newcommand\ran[1]{}
\newcommand\sv[1]{}
\newcommand\MM[1]{}
\fi



\title[Towards High End-to-end Utility of Gradient Compression]{Beyond Throughput and Compression Ratios: Towards High End-to-end Utility of Gradient Compression}
\author{Wenchen Han}
\affiliation{%
  \institution{University College London}%
  \country{}%  
}

\author{Shay Vargaftik} 
\affiliation{%
  \institution{VMware Research}%
  \country{}%  
}

\author{Michael Mitzenmacher}
\affiliation{%
  \institution{Harvard University}%
  \country{}%  
}

\author{Brad Karp}
\affiliation{%
  \institution{University College London}%
  \country{}%  
}  

\author{Ran Ben Basat}
\affiliation{%
  \institution{University College London}%
  \country{}%  
}

\acmYear{2024}\copyrightyear{2024}
\setcopyright{acmlicensed}
\acmConference[HOTNETS '24]{The 23rd ACM Workshop on Hot Topics in Networks}{November 18--19, 2024}{Irvine, CA, USA}
\acmBooktitle{The 23rd ACM Workshop on Hot Topics in Networks (HOTNETS '24), November 18--19, 2024, Irvine, CA, USA}
\acmDOI{10.1145/3696348.3696857}
\acmISBN{979-8-4007-1272-2/24/11}

\begin{abstract}
    
Gradient aggregation has long been identified as a major bottleneck in today's large-scale distributed machine learning training systems. One promising solution to mitigate such bottlenecks is gradient compression, directly reducing communicated gradient data volume. 
However, in practice, many gradient compression schemes do not achieve acceleration \mbox{of the training process while also preserving accuracy.}

In this work, we identify common issues in previous gradient compression systems and evaluation methodologies. These include excessive computational overheads; incompatibility with all-reduce; and insufficient evaluation methods, such as not using an end-to-end metric or using a 32-bit baseline instead of the stronger 16-bit baseline. 
{We revisit common compression approaches (sparsification, quantization, and low-rank decomposition) and demonstrate how considering the above issues can lead to minor but strategic design changes, resulting in notably better performance.
Our goal is to raise awareness of the need for design and evaluation standards that naturally translate to the end-to-end utility of gradient compression.}
\end{abstract}

\begin{document}

\maketitle

\vspace{-0.2cm}
\section{Introduction}

Distributed Data-Parallel (DDP) training~\cite{NIPS2012_6aca9700} is the de-facto paradigm for large-scale distributed machine learning training systems. 
A key obstacle to efficient DDP training is the large communication volume of aggregating and synchronizing the gradients~\cite{sapio2021scaling, tang2020communication, wang2023hi, wang2024towards}. 
Further, the training hardware's processing speed advances faster than the network bandwidth~\cite{communication-computation-tension},  exacerbating this issue. 

One promising direction to alleviate the communication bottleneck is to apply a gradient compression scheme~\cite{ on-the-utility, bai2021gradient, bernstein2018signsgd, fei2021efficient, kim2019parallax, li2024accelerating, thc,  m2021efficient, topk, powersgd, wang2018atomo,  wang2023hi, wang2023cupcake, terngrad,ben2020send,chen2024justintime}. 
Gradient compression aims to reduce the communicated data volume~\cite{wang2024towards}. 
Most schemes are lossy,
meaning they introduce some compression error, which the schemes 
try to minimize. Nevertheless, as pointed out by prior studies~\cite{on-the-utility, wang2023cupcake}, compression schemes often end up yielding a limited speedup in practice or compromise the model's accuracy. 

%We are therefore motivated to dig into the root causes that can result in a less satisfying performance in practice than research papers appear to claim. 
% Our goal is to uncover the issues in previous studies that lead to degraded end-to-end utility in practice.
%\revise{Our goal is to raise awareness of the need for design and evaluation standards that naturally translate to the end-to-end utility of gradient compression}
We seek to uncover the root causes that lead to degraded end-to-end utility in practice. %than research papers appear to claim.  
% \revold{To achieve this, we conduct a case study (Section~\ref{sec:case-study}), with examples of three main gradient compression types: sparsification (TopK~\cite{topk2, topk}), quantization (THC~\cite{thc}), and low-rank decomposition (PowerSGD~\cite{powersgd}).} 
%
We classify our findings into two categories, design and evaluation, and exemplify how to address these issues
via a case study (Section~\ref{sec:case-study}) that covers three main gradient compression types: sparsification (TopK~\cite{topk2, topk}), quantization (THC~\cite{thc}), and low-rank decomposition (PowerSGD~\cite{powersgd}).

\smallskip
\subp{Design.} One common issue is the compression's computational overhead~\cite{on-the-utility, xu2021grace, wang2023cupcake}. Our study finds that some components of compression incur superlinear computational complexity~\cite{qr-decomposition, hedayat1978hadamard, thc, powersgd}; others have GPU-inefficient memory access patterns~\cite{gpu-mem, qr-decomposition, hedayat1978hadamard}. Another challenge is that many compression schemes are incompatible with the all-reduce collective~\cite{collective-nccl, sanders2019sequential}, which is inherently more scalable than the all-gather collective that generates higher traffic overhead, or the parameter server aggregation~\cite{li2014scaling} that has one-to-many and many-to-one communication patterns~\cite{sapio2021scaling, shashidhara2022flextoe, vasudevan2009safe}. 
We elaborate on these particular issues in our case study (Section~\ref{sec:case-study}), and based on our consideration,  propose techniques such as Chunking, Partial Rotation, and Saturation to improve performance. 
%We argue that those techniques could potentially be adapted to other compression algorithms of this type as well.

\smallskip
\subp{Evaluation.} {Gradient compression studies (e.g.,~\cite{on-the-utility, bai2021gradient, fei2021efficient, kim2019parallax, li2024accelerating, thc, m2021efficient, wang2023hi, wang2023cupcake, ben2024optimal}) often choose the compression ratio (the amount of reduced communication volume) and throughput as their design objectives and evaluation metrics, and compare with a full precision (FP32) baseline. We argue that such an evaluation is insufficient for the following reasons. 
First, neither the throughput nor the compression ratio reflect the accuracy degradation.
In an end-to-end sense, gradient compression aims at optimizing the time it takes to reach a target accuracy, \ie, the time to accuracy (TTA). An excessively aggressive scheme to cut down the communication overhead
% (Figure~\ref{fig: topk}) 
may improve the throughput but often results in degraded TTA due to the high compression error that dominates the convergence speed. 
%Second, half-precision (FP16) is widely supported in ML hardware and FP16 compression and aggregation are found to be more performant in terms of TTA than FP32 (Figure~\ref{fig: topk}). 
Second, half-precision (FP16)~\cite{fp16} is found to be a stronger baseline. This is because it
requires half the number of bits and is widely supported in ML hardware, which can commonly make more FP16 ops per second compared to FP32~\cite{a100, fp16-better, ho2017exploiting}. Accordingly, FP16 compression and aggregation are found to be~\cite{jia2018highly} more performant in terms of TTA than FP32.
%(Figure~\ref{fig: topk}). 
The evaluation is therefore convincing only if the gradient compression scheme outperforms the higher FP16 bar. 
We refer to the TTA improvement over this FP16 baseline as a method's \emph{utility}.
We demonstrate the usefulness of the proposed evaluation methods using the TTA metric and FP16 baseline in our case study.
}

In summary, we make the following contributions:

1) We identify common design issues of gradient compression systems that restrain their training speedup in practice or compromise the model's accuracy.
%(Section~\ref{sec: issues}).

2) We identify common evaluation insufficiencies that may lead to suboptimal design choices that do not translate to improved end-to-end utility in practice. % and lead to \mbox{wrong conclusions. }

3) {We exemplify these issues in a case study covering three main types of gradient compression schemes and propose optimization techniques that address these issues resulting in better utility.}

%\revold{3) We conduct preliminary experiments to demonstrate the utility of our proposed techniques.}

% \section{Limitations of Prior Works}

%\section{Issues in Previous Research}\label{sec: issues}

\section{\!\resizebox{0.9320279302510\linewidth}{!}{{Gradient Compression~Challenges}}}\label{sec: issues}

% \Wenchen{In the following sections, we elaborate on the following three arguments. Speeding up compression; achieving compatibility to arbitrary all-reduce topology; better evaluation of gradient compression (comparing with TTA as the meaning of end-to-end for compression, FP16 as a natural lossy compression method with extensive hardware support should be compared.)}

In this section, we overview common fallacies and issues in the design and evaluation of modern gradient compression schemes. In particular, we focus on the design issues (that result in degraded end-to-end performance) associated with high computational overhead and incompatibility with all-reduce, the de facto standard for DDP. We also consider evaluation practices that insufficiently model the end-to-end utility of a gradient compression system, including the usage of metrics that do not capture end-to-end performance, and inappropriate or weak baselines. {We pick several state-of-the-art system papers of gradient compression~\cite{on-the-utility, bai2021gradient, fei2021efficient, kim2019parallax, li2024accelerating, thc, wang2023hi, wang2023cupcake} for analysis, as listed in Table~\ref{tab:paper-survey}.}

%Before diving into the details of three examples of gradient compression methods, 
%We start by presenting an overview of common issues in previous research. We first describe specific performance issues of the schemes that limit their utility, including high computational overhead and incompatibility with all-reduce. 
%We then look more generally to insufficient evaluation. \ran{unclear sentence}
%Many works, for example, mainly look at the training throughput (measured in images or tokens per second) when using compression. However, training throughput alone is not a suitable metric, as what is important is the end-to-end improvement in performance, which generally corresponds to time to a desired accuracy compared with a reasonable baseline. Accordingly, both computational costs and accuracy come into play. 
%\MM{This next sentence feels out of place and I suggest moving or removing it.}
% We note that the scope of works that we consider include three categories: sparsification-based compression~\cite{}, quantization-based compression~\cite{}, and low-rank based compression~\cite{}. %We have identified two categories of limitations: the limitations of the design of the gradient compression schemes themselves, and the insufficiency of their evaluation. % We support our claims by diving into three examples of different types of compression: TopK compression (sparsification based), THC (quantization based), and PowerSGD (low-rank based).

% \MM{Rewrote the below paragraph also}
% \subsection{Limitations in the Algorithmic Designs}
\vspace{-0.15cm}
\subsection{Design Issues}\label{subsec:perf-issue}
%\vspace{-0.1cm}
% \MM{Compression overhead is not the right term.  You mean the extra computational overhead of compression.}
% \smallskip
\subp{Computational overhead.} As shown in previous studies~\cite{on-the-utility, xu2021grace,wang2023cupcake}, gradient compression often underperforms an uncompressed baseline in terms of its training throughput.
%Gradient compression promises to reduce communication, but the price is generally additional computational overhead for the compression. Accordingly, as shown in previous studies~\cite{on-the-utility, wang2023cupcake}, many state-of-the-art compression schemes fail to outperform even the FP32 baseline (which aggregates gradients in full precision) in terms of their training throughput.
Computational overhead for compression plays a key role here. 
In this work, we identify components that create bottlenecks through excessive computation or using the GPU's global memory~\cite{gpu-mem} with inefficient memory access patterns.

%\revise{To minimize the computational overhead, gradient compression should preferably be friendly to the training hardware -- such as a contiguous memory access pattern on GPUs to avoid frequent access to the global memory~\cite{gpu-mem}.
%}
%\smallskip
%%and suggest corresponding improvements.~\Wenchen{Remark: locality issue}.
%\smallskip
\subp{Incompatibility with all-reduce.}
There are two common types of gradient collection methods: collection at a centralized parameter server (PS)~\cite{li2014scaling}, and decentralized collective operations such as all-gather and all-reduce~\cite{collective-nccl, sanders2019sequential}.
All-reduce (including ring all-reduce~\cite{ring} and tree all-reduce~\cite{sanders2009two}) is inherently more scalable than all-gather~\cite{collective-nccl, sanders2019sequential} and PS aggregation~\cite{li2014scaling}, since many-to-one communications~\cite{shashidhara2022flextoe, vasudevan2009safe} incur temporal congestion, and RDMA NICs face dropped performance maintaining too many connections~\cite{wang2023srnic}. Recently, \cite{jiang2020unified} explored the PS co-located mode that reduces the temporal load on any specific worker, but still suffers from the many-to-one and one-to-many communications.

Unfortunately, previous gradient compression algorithms are often not compatible with all-reduce collectives~\cite{on-the-utility}. 
%Unlike the parameter server (PS) architecture, where everything is aggregated in a single location, compression in an all-reduce topology is more challenging due to the multi-hop aggregation pattern. 
In all-reduce, unlike in the PS architecture, workers on an intermediate hop receive partially aggregated gradients, add their own, and then send the updated result to the next hop. The difficulty lies in avoiding decompression and recompression at each worker, as this would lead to significant computational overheads and accumulation of compression-induced errors, which could negatively impact the training process.

% \noindent\textit{Proposed principle.} \revise{Processing partially aggregated gradients should not involve decompression and recompression at each hop (while meeting communication constraints). Otherwise, this can incur significant computational overheads and accumulation of compression-induced errors that negatively impact the training.
% }
%Unfortunately, most prior gradient compression schemes lack compatibility with all-reduce collectives~\cite{on-the-utility}. 
%Unlike the PS architecture in which everything is aggregated in one place, in an all-reduce topology compression is more challenging as it needs to adapt to the multi-hop aggregation pattern. 
%In such an aggregation pattern, workers get partially aggregated results, add their own, and send the result to the next hop. The challenge is that we wish to avoid decompressing and compressing at each worker, as otherwise the computational overheads can become excessive, and any errors introduced by compression can accumulate to levels that would harm the training process.
%we need to collect and aggregate the gradients of different workers alongside a path of length $\ge 2$, ensuring that the aggregation is not computationally heavy and that some certain properties (detailed in examples in Section~\ref{}) that hold in the first hop do not break along the aggregation path. 

\iffalse
\begin{table}[]
    \centering
    \begin{tabular}{c|c}
         &  \\
         & 
    \end{tabular}
    \caption{Caption}
    \label{tab:evaluation-prior-works}
\end{table}
\fi

\begin{table*}[t]
    \centering
    \resizebox{0.85\linewidth}{!}{
\begin{tabular}{l|l|l|l|l|l|l|l|l|}
\cline{2-9}
                                                                               & \cite{on-the-utility} & \cite{bai2021gradient} & \cite{fei2021efficient} & \cite{kim2019parallax} & \cite{li2024accelerating} & \cite{thc} & \cite{wang2023hi} & \cite{wang2023cupcake} \\ \hline
\multicolumn{1}{|l|}{Comparing with the stronger FP16 baseline}             & \ding{55}             & \ding{55}              & \ding{55}               & \ding{55}              & \ding{55}                 & \ding{55}  & \ding{55}         & \ding{55}              \\ \hline
\multicolumn{1}{|l|}{Considering compression error for system design}          & N/A                                & \ding{55}              & \ding{51}               & N/A                                 & \ding{51}                 & \ding{51}  & \ding{55}         & \ding{55}              \\ \hline
\multicolumn{1}{|l|}{Evaluation on end-to-end performance (in how many tasks)} & 0/3                                & 2/8                                 & 1/6                                  & 3/4                                 & 4/4                                    & 3/7                     & 4/4                            & 3/3                                 \\ \hline
\multicolumn{1}{|l|}{Higher throughput results in better time to accuracy}     & N/A                                & \ding{51}              & \ding{51}               & \ding{51}              & \ding{55}                 & \ding{51}  & \ding{51}         & \ding{55}              \\ \hline
\multicolumn{1}{|l|}{All-reduce compatibility for new compression algorithms}  & N/A                                & N/A                                 & \ding{55}              & \ding{51}%Not Scalable               
& \ding{51}                 & \ding{55}  & N/A                            & \ding{55}              \\ \hline
\end{tabular}
    % \begin{tabular}{|l||l|l|l|l|l|l|l|l|} \hline
    %      & \cite{on-the-utility} & \cite{bai2021gradient} & \cite{fei2021efficient}   & \cite{kim2019parallax} &   \cite{li2024accelerating} & \cite{thc}  & \cite{wang2023hi} & \cite{wang2023cupcake} \\ \hline\hline
    %     Comparing with an the stronger FP16 baseline & \ding{55} & \ding{55} & \ding{55} & \ding{55} & \ding{55} & \ding{55} & \ding{55} & \ding{55} \\ \hline
    %     Considering compression error in addition to throughput for their design & N/A & \ding{55} & \ding{51} & N/A & \ding{51} & \ding{51} & \ding{55} & \ding{55} \\ \hline
    %     Evaluation on end-to-end performance (in how many tasks) & 0/3 & 2/8 & 1/6 & 3/4 & 4/4 & 3/7 & 4/4 & 3/3 \\ \hline
    %     %Throughput end-to-end agreement in all tasks 
    %     Higher throughput results in better time to accuracy
    %     & N/A & \ding{51} & \ding{51}  & \ding{51} & \ding{55} & \ding{51} & \ding{51} & \ding{55} \\ \hline
    %     All-reduce compatibility for new compression algorithms & N/A & N/A & \ding{55} & N/A & \ding{51} & \ding{55} & N/A & \ding{55} \\ \hline        
    % \end{tabular}
    }
    \vspace{.1cm}
    \caption{{Assessment of prior gradient compression systems. Here, `N/A' means that the criterion is not applicable. %\cite{kim2019parallax} supports all-reduce but incurs $n\times$ communication where $n$ is the number of workers. %Throughput end-to-end agreement means the proposed solution achieves both higher throughput and end-to-end performance than baselines. 
    }}
    \vspace{-0.5cm}
    \label{tab:paper-survey}
\end{table*}

\vspace{-0.1cm}
\subsection{Evaluation Issues}\label{subsec:evaluation-issue}
%\vspace{-0.1cm}

%We argue that previous evaluations of many state-of-the-art gradient compression methods~\cite{thc, powersgd, fei2021efficient, bai2021gradient} do not appropriately evaluate the utility of gradient compression schemes.  
%We accordingly suggest better evaluation methods, emphasizing two key aspects.

% THC: 1 TTA figure, 4 throughput/time
% on-the-utility: 0 TTA figures, 7 throughput/time
% HiPress: 1 TTA figure, 4 throughput/time
% OmniReduce: 0  TTA figures
% Espresso: 1 TTA figure, 3 throughput/time

\subp{The choice of an end-to-end metric.} 
Prior works~\cite{on-the-utility, bai2021gradient, fei2021efficient, kim2019parallax, li2024accelerating, thc,  wang2023hi, wang2023cupcake} primarily focus on optimizing the {\em training throughput} metric. Admittedly, training throughput can be a cheap metric to evaluate on, as it does not require a full training process. Also, training throughput is a suitable metric for lossless compression, where compression error need not be taken into account.  However, with lossy compression, training throughput is not an end-to-end metric, as it does not consider the impact of compression error on the model's  accuracy~\footnote{Here accuracy refers to the goal metric of the trained model, such as classification accuracy in recognition tasks and perplexity (the model's confidence in predicting the correct next token) for language modeling.}. As shown in Table~\ref{tab:paper-survey}, several prior gradient compression systems do not take into account compression error in their design, do not include end-to-end performance evaluation, or do not achieve acceleration compared to the baselines in terms of time-to-accuracy.

In an end-to-end sense, the utility of a gradient compression scheme is exactly the training time saved to train a model.
Accordingly, we propose that the time to accuracy (TTA) should be the main end-to-end metric. 
%
% Here, `accuracy' refers to the goal metric of the trained model, e.g., classification accuracy in recognition tasks or perplexity (the model's confidence in predicting the correct next token) for language modeling tasks.
%
Critically, TTA is a \emph{2-dimensional} metric in which for each accuracy target, an algorithm has a measured training time required to meet that target.
One challenge is that each compression scheme is then represented by a \emph{curve}, not a single data point, and the curves can intersect, so that which scheme is better can depend on the setting. We argue here that research papers, in particular, should present \emph{TTA curves} when comparing compression schemes, as otherwise, they are leaving out the key information for understanding comparative performance.

While TTA curves still leave room for interpretation, discussing the curves provides a better means of comparing schemes than arbitrarily chosen times or accuracies (particularly when selected to make a new solution appear strong).
In most cases, we expect the focus to be on accuracies close to the accuracy attained by an uncompressed baseline (which is typically desired). 
But in practice, as we later verify empirically, not all compression schemes can meet all accuracy targets, as compression often results in lower final model accuracy than the uncompressed baseline.
Typically, for research comparing different schemes, we suggest that different techniques be run until convergence (\eg, according to an early stopping standard~\cite{prechelt2002early}) to produce the TTA curves.

End-to-end evaluation is inevitably costly but necessary for gradient compression research, as focusing solely on metrics that are not end-to-end (such as training throughput) results in an insufficient understanding of the system performance and incomplete conclusions. However, for tasks such as parameter tuning, there are useful cheaper proxies that (combined with training throughput) quickly deliver a rough idea of the convergence speed at the beginning of the training. One example is vector's normalized mean squared error (vNMSE)~\cite{karimireddy2019error, thc, vargaftik2021drive, vargaftik2022eden,benaccelerating}, which measures the compression error between the true gradients' average and its estimate from the compressed gradients.

We note that some other end-to-end metrics could be a better fit in certain circumstances, such as cost-to-accuracy and power-to-accuracy. There may be cases where one is willing to use a less robust metric, such as by picking a specific target accuracy (e.g., $95\%$), but this is generally not a suitable goal when comparing compression schemes generally (as in research), and should be justified when done.
\begin{table}[!t]
    \centering
    \resizebox{1\linewidth}{!}{
    \begin{tabular}{|c||c|c||c|c|} \hline
        Task & TF32+FP16 & TF32+FP32 & FP32+FP16 & FP32+FP32\\ \hline
        BERT & $3.32$  & $2.44$ & $3.17$ & $2.36$ \\ \hline
        VGG19 & $9.31$  & $6.59$ & $8.73$ & $6.37$ \\ \hline
    \end{tabular}
    }
    \vspace{.02272cm}
    \caption{Throughput (in rounds per second) of baselines varying training and communication precision.  FP32+FP16 means we train with FP32 and communicate in FP16 precision.
    TF32, FP32, FP16 refer to Tensorfloat \protect\cite{tensorfloat}, full- and half-precision.}
    \label{tab:bl-throughput}
    \vspace{-1.13cm}
\end{table}

%\smallskip
\subp{The choice of baselines.} Most of the literature (\cite{bai2021gradient, on-the-utility, fei2021efficient,kim2019parallax, thc, m2021efficient, powersgd, wang2023hi, wang2023cupcake})
%\Wenchen{Done: "ALL" citations that I find choose fp32 as the baseline.} 
compares against a baseline using full precision (FP32) aggregation. However, the use of half-precision (FP16)~\cite{fp16}, which has gained wide hardware support~\cite{fp16-better, ho2017exploiting}, provides a stronger baseline. 
It achieves $50\%$ less communication and thus much higher throughput (see Table~\ref{tab:bl-throughput}), and as shown in our experiments, the accuracy degradation is negligible. 
%\revise{A compression scheme is still of low utility if it does not perform better than the stronger FP16 baseline.}
%\revise{\textit{In principle}, a compression scheme is of high utility only if it outperforms a strong baseline like FP16 that is simple (intrinsically supported by hardware) yet faster and accuracy-preserving.}

\vspace{-0.13cm}
\section{A Case Study}\label{sec:case-study}
%\vspace{-0.1cm}
In this section, we conduct a case study to better elaborate on the design and evaluation issues of gradient compression. 
We focus on three common categories of gradient compression method: sparsification~\cite{topk2, kim2019parallax, li2024accelerating, topk} (\S\ref{subsec:topk}), quantization~\cite{alistarh2017qsgd,bernstein2018signsgd, thc, terngrad} (\S\ref{subsec:thc}) and low-rank decomposition~\cite{powersgd, wang2018atomo} (\S\ref{subsec:powersgd}). Namely, as representatives, we use TopK sparsification~\cite{topk2, topk}, THC quantization~\cite{thc} and PowerSGD decomposition~\cite{powersgd}. 
%We also discuss how our optimizations \mbox{can apply to other algorithms.}
%
We also introduce strategic design changes to these techniques, showing how considering issues motivates better design of gradient compression.% \Wenchen{"Strategic design changes" change to "strategic design and evaluation changes" to these schemes.}

%respectively as representative compression examples to each category, but will also elaborate on how our experimental findings and proposed optimizations for each example could be potentially applied to other algorithms for this category as well.

%\subp{Common evaluation setups.} 
\looseness=-1
Our prototypes are implemented in NCCL~\cite{nccl} and PyTorch DDP~\cite{li2020pytorch}. We choose two common tasks for evaluation, namely VGG19~\cite{simonyan2014very} for classification on TinyImageNet~\cite{le2015tiny} and BERT-large~\cite{devlin2018bert} for masked language modeling~\cite{salazar2019masked} on WikiText-103~\cite{wikitext}. The per-worker batch size is set as $32$ and $4$, respectively.
%We use the time to perplexity/top-1 accuracy as the end-to-end metric but also include the throughput results to analyze solely the training speedup of compression. 
The prototypes are deployed on a testbed with two nodes, each equipped with two NVIDIA A100 GPUs~\cite{a100} (for a total of 4 GPUs) and a Mellanox ConnectX-6 100Gbps NIC~\cite{connextx6}. Early stopping~\cite{prechelt2002early} is applied {to terminate training upon model convergence. We record the testing perplexity for BERT-large and the top-1 accuracy for VGG19. To plot the TTA figures, we apply a rolling average over $3750$ rounds (0.3 epochs) for BERT-large { and $7810$ rounds (10~epochs)~for~VGG19.} }\\
%\Wenchen{Alternatively: 0.3 epochs for BERT-large and 10 epochs for VGG19.} % All the models are trained until convergence (\eg, according to an early stopping standard~\cite{prechelt2002early}) 

We measure the communication overhead in terms of the all-reduce input size, in bits per coordinate denoted by $b$. Note that the actual bandwidth required depends on the specific all-reduce scheme. In ring-all-reduce, the total overhead is roughly $2 b$ per coordinate (reduce-scatter and all-gather). 

\vspace{-0.1cm}
\subsection{TopK Sparsification}\label{subsec:topk}

\begin{figure*}[t]
        \vspace{-0.03cm}
        \centering
        \begin{minipage}[t]{0.5\linewidth}{
		\vspace{-0.00in}
		\begin{center}
		\includegraphics[width=\textwidth, ]{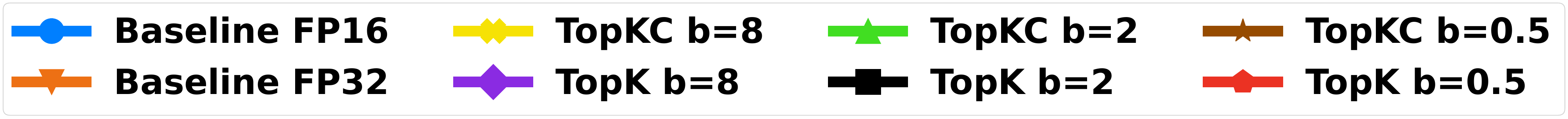}
		\end{center}
		}
        \end{minipage}

        \vspace{-0.23cm}
	\centering
        \subfigure[BERT-large]{
		\begin{minipage}[t]{0.44\linewidth}{
		\vspace{-0.00in}
		\begin{center}
		\includegraphics[width=\textwidth, ]{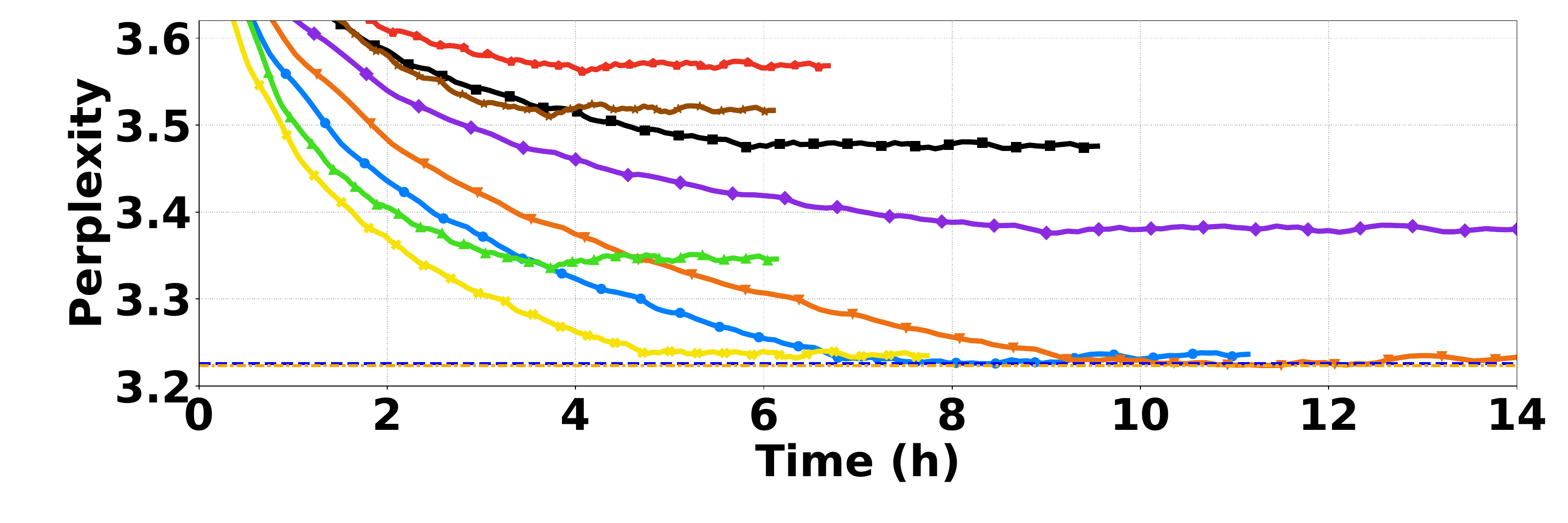}
		\end{center}
		}
            \vspace{-2.3mm}
		\label{subfig:topk-lm}
		\end{minipage}
	}
	%
        % \vspace{-0.3cm}
        % \centering
	\subfigure[VGG19]{
		\begin{minipage}[t]{0.44\linewidth}{
		\vspace{-0.00in}
		\begin{center}
		\includegraphics[width=\textwidth, ]{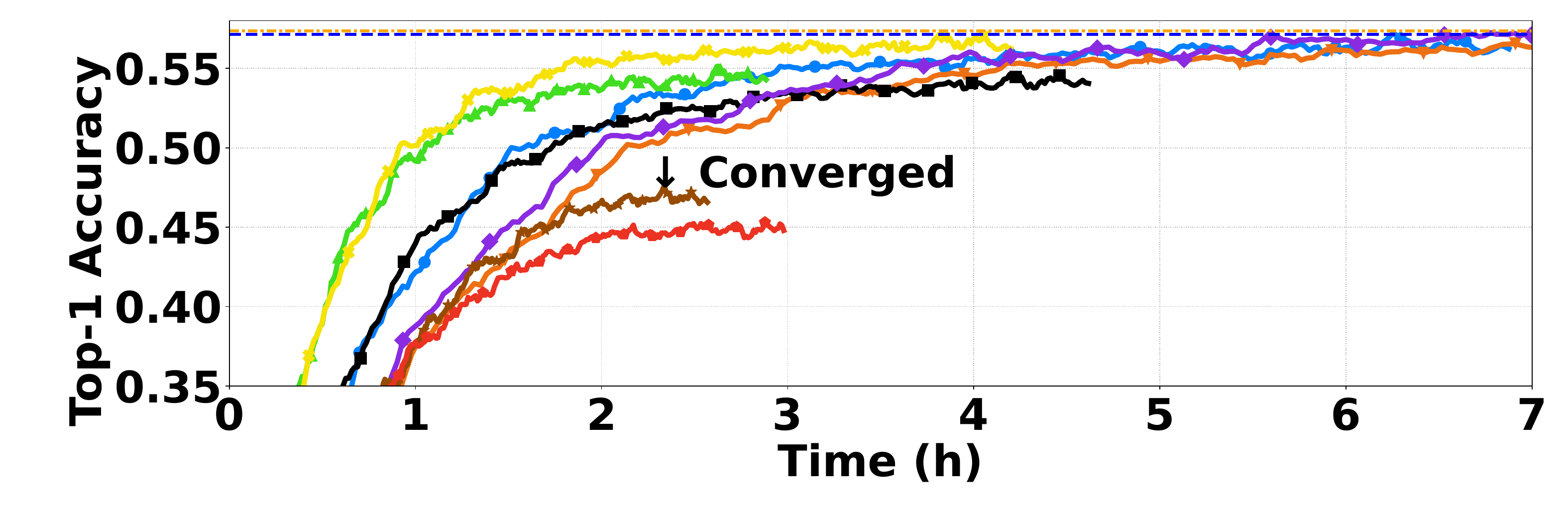}
		\end{center}
		}
        \vspace{-2.3mm}
		\label{subfig: topk-image}
		\end{minipage}
	}
	\vspace{-0.35cm}
	\caption{The TTA (rolling averaged) of our TopK Chunked (TopKC) solution compared with TopK and the baselines. The dashed lines indicate the converged perplexity/accuracy for Baseline FP16 and Baseline FP32 respectively.  The training of each method stops after a given number of epochs (and not hours) after convergence.}
        \vspace{-0.15cm}
	\label{fig: topk}
	
\end{figure*}

\subsubsection{Background and issues}\label{subsubsec:topk-background}\quad

Sparsification compresses by only sending some of the gradient coordinates. 
Intuitively, one would want to send the top $K$ coordinates with the largest \emph{aggregated} (summed across gradients) value, which we term Global TopK. Unfortunately, finding these indices is challenging without computing the aggregated gradient to begin with. Thus, in practical (local) TopK sparsification (e.g.,~\cite{topk2, topk3, topk}) implementations, each worker extracts its $K$ largest (in absolute value) coordinates and their indices to be aggregated later. Prior works typically use TopK compression with the less scalable PS architecture or a distributed all-gather collective~\cite{collective-nccl, sanders2019sequential} as their aggregation scheme but do not implement it with all-reduce.

\begin{table}[t]
    \centering
    \resizebox{1\linewidth}{!}{
    \begin{tabular}{|c||l|} \hline
        \textbf{Notation} & \textbf{Meaning} \\ \hline \hline
        $b$ & Volume of all-reduce communication in bits per coordinate. \\ \hline
        $n$ & The number of workers. \\ \hline
        $d$ & The total number of coordinates in each gradient vector. \\ \hline\hline
        $K$ & The number of top $K$ coordinates selected to be aggregated. \\ \hline
        $C$ & The chunk size of TopKC which gradients are partitioned into. \\ \hline
        $J$ & The number of top chunks selected for aggregation in TopKC. \\ \hline
        $J'$ & The total number of coordinates in top $J$ chunks, \ie, $J'=JC$. \\ \hline\hline
        % $c_p^{(i)}$ & The $p$th chunk for the $i$th worker. \\ \hline\hline
        $q$ & The $q$-bit integers into which each FP32 gradient is quantized. \\ \hline
        $l$ & Number of RHT iterations. Gradients are padded to size $2^l$. \\ \hline
        $l'$ & The number of iterators for partial rotation. $l' \le l$. \\ \hline
        $Sat(\cdot, \cdot)$ & The saturation operator. \\ \hline\hline
        $r$ & The target matrix rank $r$ for PowerSGD. \\ \hline

    \end{tabular}
    }
    
    \vspace{.1cm}
    \caption{Notation used in our case study.}
    \vspace{-1cm}
    \label{tab:notations}
\end{table}

%\smallskip
\subp{Computational overhead.} The TopK selection operation and rearrangement of coordinates is a major bottleneck~\cite{shanbhag2018efficient} (see \S\ref{subsubsec:topk-evaluation}). The reason is that they involve non-consecutive memory accesses with poor locality, which slows down the processing speed of the GPU~\cite{gpu-mem}. 
%In fact, our profiling (Table~\ref{tab:topk-throughput}) shows that these two operations can slow down the training throughput by up to $9.8\%$ and $14.3\%$ for BERT-large and VGG19 respectively.

%\smallskip
\subp{All-reduce incompatibility.} 
With a standard all-reduce operation without compression with $n$ workers and gradients of size $d$, each worker sends and receives $d/n$-sized blocks at each of $2(n-1)$ steps, and these blocks are summed (reduced) at intermediate workers.
When implementing TopK in a distributed setting, a challenge is that the coordinates sent by each worker may be different, resulting in up to $nK$ distinct coordinates, which may not give significant compression if $nK$ is comparable to or even larger than $d$.

\subsubsection{Our proposed improvement:  Chunks}\label{subsubsec:topk-solution} \quad

\subp{Overview.} %\Wenchen{TODO: there's a need to distinguish between our chunk-based "global TopK" (aligned version) adapted for all-reduce, and "block-based" local topk proposed by Omnireduce (which adapts only for PS not all-reduce, since they are not "coordinates/blocks aligned". They don't have our first step, so they aggregate local topk instead of global topk.).} 
We propose a TopK variant dubbed TopK Chunked (TopKC in short) that has less computational overhead and is compatible with all-reduce to approximate Global TopK. The key idea is to perform a lightweight step to reach consensus on a good set of coordinates to aggregate (that are not the top $K$ coordinates). 
%In return, consensus will not strictly be the global TopK coordinates, but instead achieve a good approximation to them. 
%\MM{I think our approximation is in the heuristic, no provable sense;  make the clear?}
%Specifically, we propose computing the TopK chunks of the gradient vector.  
In TopKC, for a hyperparameter $C$, each worker partitions its gradient into $\lceil d/C \rceil$ fixed-sized chunks. The workers then decide which chunks to communicate with an initial all-reduce communication round. In this round, each worker sends the squared norm of each chunk, allowing all workers to agree on the $J<d/C$ chunks with the largest sum of L2 norms. Intuitively, chunks with large sum-of-norms are likely to contain large coordinates that are worth aggregating.
%and weigh each chunk by its squared L2 norm.
%%\Wenchen{I prefer the word "feature". Squared L2 norm is just one example of features. Other potential feature may be an ams sketch.}  
%It then requires little communication to reach a consensus of the global TopK chunks, and 
We then aggregate these agreed-upon chunks using all-reduce, thus making TopKC all-reduce-compatible. A benefit of TopKC is that our memory access pattern is mainly sequential and the expensive top-$K$ calculation~\cite{shanbhag2018efficient} operates on fewer values than TopK ($\lceil d/C \rceil$ rather than $d$), allowing faster execution in practice.
%\Wenchen{To discuss: Added the argument that TopK calculation is expensive.}

%Beyond providing all-reduce compatibility, there are additional benefits. The overheads are reduced since the selection of chunks requires less computation than the selection of individual coordinates, and memory accesses on chunked data are sequential and thus GPU-friendly. 
%2) Given a fixed $J$, the number of bits per coordinate required $b$ is reduced, since there is no longer needs to send the $32$-bit indices.

\begin{table}[t]
    \centering
    \resizebox{0.7\linewidth}{!}{
    \begin{tabular}{|c||c|c|c|} \hline
        Compression & $b=0.5$ & $b=2$ & $b=8$  \\ \hline \hline
        TopKC &$0.273$ & $0.142$ & $0.0280$ \\ \hline
        TopKC Permutation & $0.398$ & $0.297$ & $0.123$ \\ \hline
    \end{tabular}
    }
    \vspace{.1cm}
    \caption{vNMSE of the aggregated gradients with TopKC and TopKC with random permutation for BERT. %The bits-per-coordinate $b$ ranges from $0.5$ to $8$.
    }
    \vspace{-1.2cm}
    \label{tab:topk-chunk-random}
\end{table}

%Admittedly, the coarse-grained TopK chunks can include coordinates that are of low significance. 
%However, w
We show that TopKC can be an effective heuristic, in part because of \textit{spatial locality}, meaning large coordinates tend to appear closer to each other. 
%With spatial locality, the TopK chunks with respect to the squared L2 norm cover most of the coordinates with the largest magnitude. \MM{The below description doesn't make sense to me.  What exactly are we trying to show here?} 
To demonstrate spatial locality, we compare against a variant, TopKC Permutation, that randomly permutes the coordinates, eliminating any spatial locality. We choose to use vNMSE~\cite{vargaftik2021drive} (Section~\ref{subsec:evaluation-issue}) to study the compression error, and follow the experimental settings shown in Section~\ref{subsubsec:topk-evaluation}.
Table~\ref{tab:topk-chunk-random} indicates that TopKC takes advantage of spatial locality, since it performs significantly better than when the coordinates are randomly permuted.  

%\smallskip
\subp{Design changes.} TopKC consists of the following steps.

1) We partition each local gradient into smaller chunks of size $C$. 
    Let $c_{p}^{(i)}$ be the $p$th chunk for the $i$th worker, so each has a vector of values $c_1^{(i)} \cdots c_{\lceil d/C \rceil}^{(i)}$. Each worker then calculates the squared L2 norm of each chunk, $||c_{p}^{(i)}||_2^2$. The squared L2 norms (in half-precision) %(in bfloat16~\cite{bfloat16}) 
    are then aggregated across workers for each chunk using all-reduce, \ie, $s_p := \sum_i ||c_p^{(i)}||_2^2$. 
    
    2) The consensus TopK chunks with the highest $s_p$ values are determined locally by all workers. These will be the global TopK chunks. The sparsified gradients containing the global TopK chunks are summed via an all-reduce communication in \textit{half-precision} (i.e., FP16).

The first stage of determining the local squared L2 norms can be computed efficiently, and all-reducing the squared L2 requires communication of $16 / C$ bits per coordinate. Adding the second stage of all-reduce communication, the total cost in bits per coordinate is  $b=16 (JC/d + 1 / C)$, where $J$ denotes the number of TopK chunks selected each round.

We believe our chunk-based aggregation approach, which allows us to cheaply coordinate the coordinates for sparsification, may be generalizable to other schemes
\cite{kim2019parallax, mstopk}.

% \subp{Discussion.} In our proposed solution, we can conclude that the idea of \textit{chunk-based} aggregation for sparsification-based enables us to achieve all-reduce compatibility, in that it allows the communication-cheap coordinate-aligned aggregation by operating on the chunk features such as squared L2 norm, which helps to reach a consensus. We claim that this technique can potentially be generalized to other sparsification-based compression schemes such as~\cite{kim2019parallax, mstopk}.

\begin{figure*}[t]
        \vspace{-0.cm}
        \centering
        \begin{minipage}[t]{0.65\linewidth}{
		\vspace{-0.00in}
		\begin{center}
		\includegraphics[width=\textwidth, ]{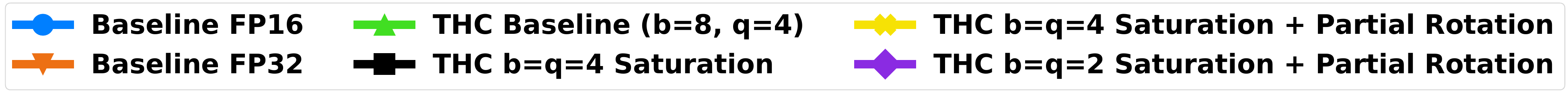}
		\end{center}
		}
        \end{minipage}
        
	\vspace{-0.23cm}

        \centering
	\subfigure[BERT-large]{
		\begin{minipage}[t]{0.44\linewidth}{
		\vspace{-0.00in}
		\begin{center}
		\includegraphics[width=\textwidth, ]{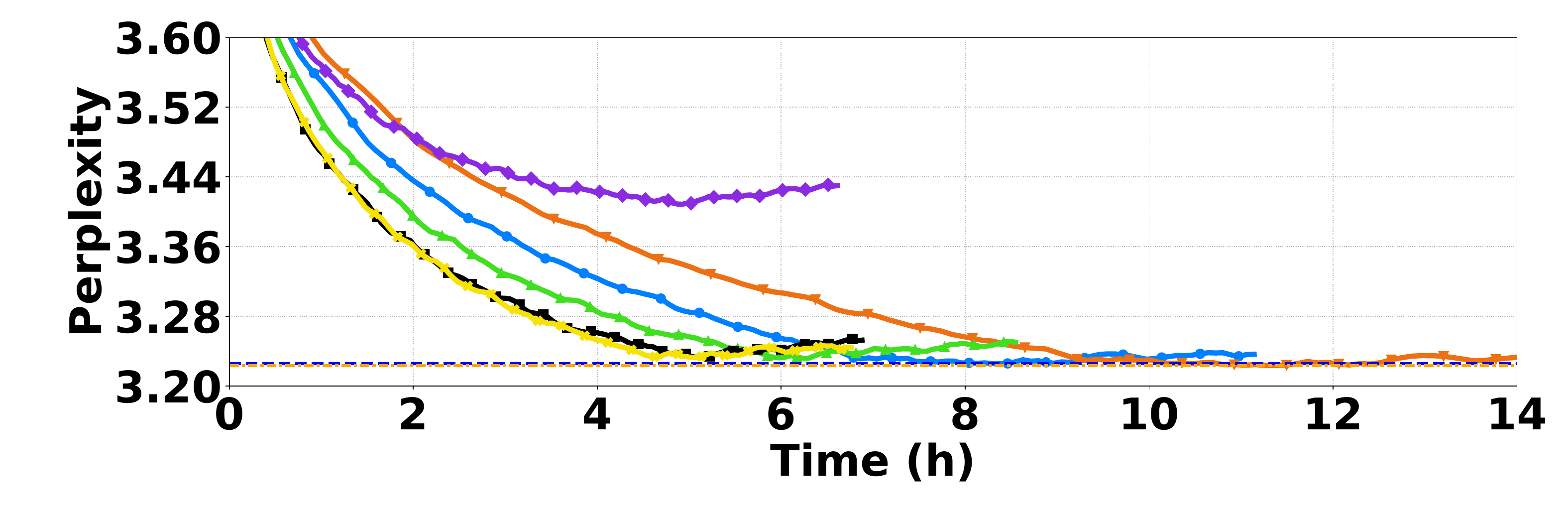}
		\end{center}
		}
            \vspace{-2.5mm}
		\label{subfig:topk-lm2}
		\end{minipage}
	}
	%
        % \vspace{-3mm}
	\subfigure[VGG19]{
		\begin{minipage}[t]{0.44\linewidth}{
		\vspace{-0.00in}
		\begin{center}
		\includegraphics[width=\textwidth, ]{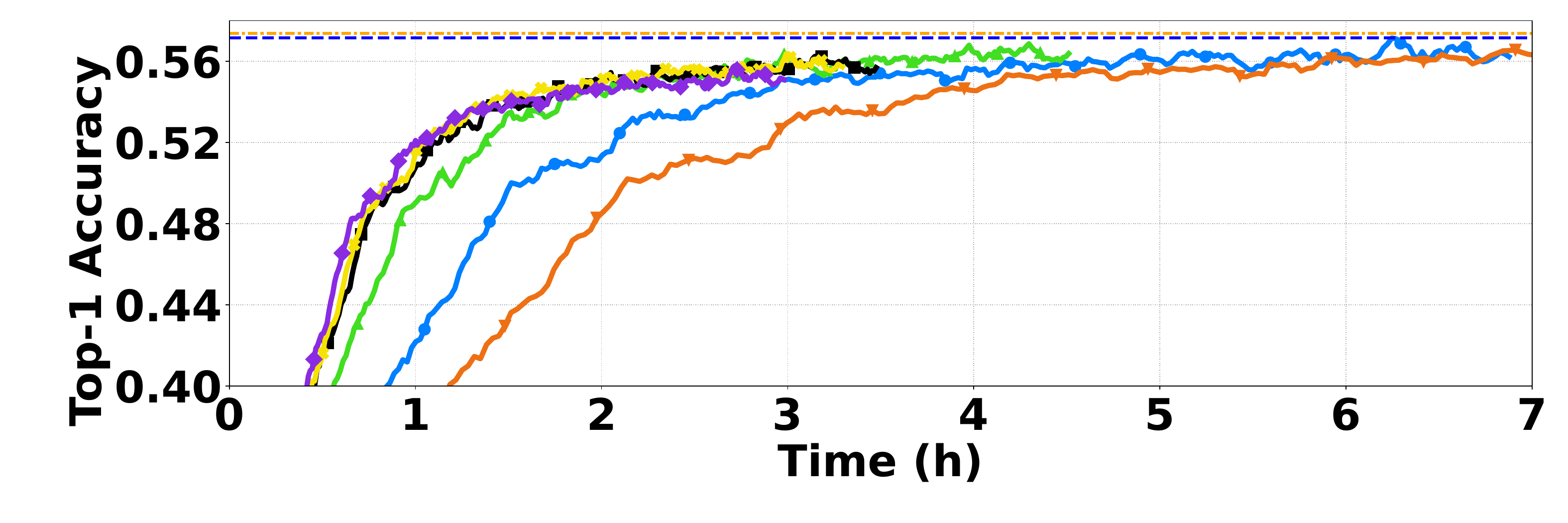}
		\end{center}
		}
            \vspace{-2.5mm}
		\label{subfig: topk-image2}
		\end{minipage}
	}
	\vspace{-0.6cm}
	\caption{The TTA of THC's simple adaptation to all-reduce compared with THC adding saturation and partial rotation.}
        \vspace{-0.2cm}
	\label{fig: thc}
	
\end{figure*}

\subsubsection{Preliminary evaluation}\label{subsubsec:topk-evaluation}\quad

\begin{table}[t]
    \centering
    \resizebox{0.8\linewidth}{!}{
    \begin{tabular}{|c|c||c|c|c|} \hline
        Task & Compression & $b=0.5$ & $b=2$ & $b=8$ \\ \hline
        \multirow{2}{*}{\begin{tabular}[c]{@{}c@{}}BERT-large \\ (345M params)\end{tabular}} & TopK & $5.53$ & $3.87$ & $2.50$\\ \hhline{~----}
         & TopKC &  $6.06$ & $6.02$ & $4.78$ \\ \hline 
         \multirow{2}{*}{\begin{tabular}[c]{@{}c@{}}VGG19 \\ (144M params)\end{tabular}} & TopK & $21.5$ & $13.9$ &  $7.60$\\ \hhline{~----}
         & TopKC &  $24.9$ & $ 22.2$ & $15.2$ \\ \hline
    \end{tabular} 
    }
    \vspace{.1cm}
    \caption{Throughput (in rounds per second) comparing the original TopK against TopK Chunked (TopKC). $b$ refers to the bits-per-coordinate as the aggregation input of the all-gather collective for TopK and all-reduce for TopKC. Note that the final accuracy is reflected in TTA curves of Figure~\ref{fig: topk}.}
    \label{tab:topk-throughput}
    \vspace{-0.4cm}
\end{table}

\begin{table}[t]
\centering
\resizebox{0.5027\linewidth}{!}{
\begin{tabular}{|c||c|c|c|} \hline
Task & $b=0.5$ & $b=2$ & $b=8$ \\ \hline
BERT & $9.7\%$ & $12.5\%$ & $8.7\%$ \\ \hline
VGG19 & $11.9\%$ & $12.1\%$ & $8.2\%$ \\ \hline
\end{tabular}
}
\vspace{-0mm}
\vspace{.1cm}
\caption{The compression overhead (the percentage of time spent on the computationally heavy components) of TopK.}
\vspace{-0.7cm}
\label{tab:topk-overhead}
\end{table}

\subp{Setup.} For both TopK and TopKC, we keep the bits-per-coordinate $b$ the same, varying from $b=0.5$ to $8$. For TopK, we follow its typical implementations~\cite{jiang2020unified, shi2019distributed} and transmit FP16-compressed values and their 32-bit indices of the top $K$ coordinates. \footnote{We note that it is possible to use 16-bit indices, e.g., by using delta-encoding and including additional coordinates to ensure the differences between consecutive indices are representable with 16 bits each. However, its GPU-unfriendly computation means that the TTA may not improve, and this does not seem to \mbox{be how TopK is implemented in practice.}}
Thus, we need to send $b=(48K)/d$ bits per coordinate. 
An all-gather collective is chosen for aggregation. For TopKC, we set $C=64$ for $b=8$ and $b=2$, and $C=128$ for $b=0.5$. % to balance between the extra communication and computation overhead and the resulting NMSE. 
% This bounds the communication overhead of the first stage within a small proportion while minimizing the NMSE. 
Denoting by $J' = JC$ the total number of coordinates that belong to one of the top $J$ chunks, $J' > K$ holds given the same $b$, thanks to the fact that TopKC saves much communication on exchanging indices for the above parameters. Finally, error-feedback~\cite{karimireddy2019error,error-feedback} is \mbox{applied to both TopK and TopKC.} 

\begin{table}[t]
    \vspace{-0.1cm}
    \centering
    \resizebox{0.55\linewidth}{!}{
    \begin{tabular}{|c||c|c|c|} \hline
        Compression & $b=0.5$ & $b=2$ & $b=8$  \\ \hline \hline
        TopK & $0.303$ & $0.185$ & $0.0865$ \\ \hline
        TopKC &$0.273$ & $0.142$ & $0.0280$ \\ \hline
    \end{tabular}
    }
    \vspace{.1cm}
    \caption{vNMSE of aggregated gradients comparing TopKC and TopK for BERT with respect to bits per coordinate $b$.}
    \label{tab:topk-chunk-approximation}
    \vspace{-.6791cm}
\end{table}

\smallskip
\subp{Evaluation issues.} First, as illustrated in Figure~\ref{fig: topk} and Table~\ref{tab:bl-throughput}, FP16 appears as a stronger baseline in terms of both TTA and throughput. It is important to compare with the stronger FP16 baseline, as TopK $b=8$ marginally outperforms FP32 in VGG19 but falls behind FP16's TTA.

Second, we see that training throughput can be a misleading metric for end-to-end performance. For example, for both TopKC and TopK, reducing $b$ from $8$ to $0.5$ enhances training throughput in BERT-large but leads to degraded TTA and final accuracy. This difference can be attributed to the increased compression error, as displayed in Table~\ref{tab:topk-chunk-approximation}. %Instead, we should choose compression solutions to target higher end-to-end performance.
% increases error and degrades performance; however, using \(b=8\) achieves the fastest convergence with the lowest throughput, contrasting with the more challenging FP16 baseline for TTA comparison.}

% {As illustrated in Figure, TopKC $b=8$ is the only solution that is favorable to Baseline FP16 in terms of TTA reaching comparable final accuracy. On the other hand, TopK solutions do not outperform the FP16 baseline for both tasks and even the FP32 baseline for BERT-large. } 
\smallskip
\subp{Analysis of our design changes.} {As shown in Figure~\ref{fig: topk}, our TopKC variation has better TTA results than TopK, and this is due to both higher throughput and better accuracy.
TopKC's throughput outperforms the original TopK by up to $2\times$ (Table~\ref{tab:topk-throughput}). The improvement can be primarily attributed to TopKC's compatibility with all-reduce, which is more communication-efficient than all-gather. 
While TopKC's computational overhead is negligible, 
even with minimal communication overhead (e.g., $b=0.5$), 
%TopK's computation still impacts the training. 
Table~\ref{tab:topk-overhead} shows that TopK's computation takes $\sim 10\%$ of the training time. %, while the .
In terms of the compression error, as indicated in Table~\ref{tab:topk-chunk-approximation}, TopKC outperforms TopK in terms of vNMSE. This can be attributed to TopKC aggregating more coordinates than TopK given the same $b$, \ie, $J'>K$.}

%\smallskip

%\subp{Discussions.} Our experimental results in Figure~\ref{fig: topk} and Table~\ref{tab:topk-throughput} also provide evidence of the benefit of choosing TTA as the main metric and FP16 aggregation as the baseline. For the former, we see that decreasing $b$ to $0.5$ enhances the training throughput, but the associated increase in error (as suggested in Table~\ref{tab:topk-chunk-approximation}) severely
%degrades performance, causing both a lower TTA curve and poor final results at convergence. Indeed, for TopKC, the experiment with $b=8$ converges the fastest on BERT-large despite having the lowest throughput among the three configurations. We also observe that Baseline FP16 poses a higher TTA bar to compare against. % The TopK with $b=2$ could reach a better TTA than Baseline FP32 but could not for Baseline FP16. 

%\vspace{-0.2cm}
\subsection{THC Quantization}\label{subsec:thc}
%\vspace{-0.1cm}

\subsubsection{Background and issues}\label{subsubsec:thc-backgrounds}\quad

\begin{table}[t]
    \centering
    \resizebox{0.88\linewidth}{!}{
    \begin{tabular}{|c|c||c|c|c|} \hline
        Task & \#bits & Full Rotation & Partial Rotation & No Rotation\\ \hline
        \multirow{3}{*}{BERT} & Sat, \(b=q=2\) & $5.59$ & $5.75$ & $5.84$\\ \hhline{~----}
         & Sat, \(b=q=4\) & $5.37$ & $5.47$ & $5.54$ \\ \hhline{~----}
         & BL \(b=8,q=4\) & $4.32$ & N/A & N/A \\ \hline

        \multirow{3}{*}{VGG19} & Sat, \(b=q=2\) & $19.9$ & $21.5$ & $22.7$ \\ \hhline{~----}
         & Sat, \(b=q=4\) & $18.4$  & $19.4$ & $20.3$ \\ \hhline{~----}
         & BL \(b=8,q=4\) & $14.2$ & N/A &  N/A \\ \hline
    \end{tabular}
    }
    \vspace{0.1cm}
    \caption{Throughput of THC with Saturation (Sat) compared with the baseline (BL) which adds $4$ more communication bits to prevent overflows during aggregation (so $b=8$). Note that the final accuracy is reflected in TTA curves of Figure~\ref{fig: thc}.}
    \label{tab:thc-throughput}
    \vspace{-1.05cm}
\end{table}

\iffalse
\begin{figure}
    \centering
    \includegraphics[width=0.9\linewidth]{partial_rotation.pdf}
    \caption{An overview of our proposed optimization of THC. Our contribution is highlighted in bold.\Wenchen{To be removed.}}
    \label{fig:overview-improved-THC}
\end{figure}
\fi

THC~\cite{thc} is a quantization-based compression algorithm designed specifically for PS architectures, where the PS can optionally be offloaded to programmable switches~\cite{Tofino, p416}. It adopts stochastic quantization to map floating-point gradients into $q$-bit integers $[0, 2^{q}-1]$. The value range between the minimum and maximum gradient values is equally split into subranges with quantized values at the boundaries, and each coordinate is stochastically rounded to one of the two nearest quantized values. One key optimization for THC to enhance quantization accuracy is to adopt the 
Randomized Hadamard Transform (RHT)~\cite{hedayat1978hadamard}, which rotates the gradient randomly before quantization and thus decreases the range between the minimum and maximum gradient values. 
% To be specific, as shown in Figure~\ref{fig:overview-improved-THC}, the RHT $H_{l}$ for a vector of size $2^{l}$ is defined as $l$ recursive steps, and the $k$th step constructs $H_k$ rotation via $H_1 \ \otimes H_{k-1}$, which operates on two neighboring $H_{k-1}$ rotated vectors of size $2^{k-1}$.

% \smallskip
\subp{Computational overhead.} We find that RHT incurs considerable overhead. As shown in Table~\ref{tab:thc-throughput}, THC with RHT is $4.4\%$ and $13.2\%$ lower in training throughput than THC without a rotation for BERT and VGG19 tasks respectively, suggesting 
improvement is possible 
if RHT can be made more efficient without a significant loss of accuracy.  The high computational overhead arises because RHT requires $O(d \log d)$ steps to compute and can have poor locality of memory accesses. 
The latter is because RHT involves memory-distant operations; for large $d$ it could not fit into the fast but small shared memory of GPUs, and \mbox{fallbacks to the slower global memory.}

% \smallskip
\subp{All-reduce incompatibility.}\label{subp:thc-issue-all-reduce} The aggregation process of a coordinate $p$ can be formulated as summing up quantized values in $[0, 2^{q}-1]$ from $n$ workers, \ie, $\sum_{i} g_{p}^{(i)}$ where $g_{p}^{(i)}$ denotes the quantized value of coordinate $p$ for worker $i$. If we allocate $b=q$ bits per coordinate for aggregation, the summed integer values could overflow the value range that $q$ bits can represent. This is not an issue for the PS architecture, where the PS is the destination for a full aggregation and can simply allocate more bits on the server to prevent overflows. For all-reduce, however, an intermediate hop has to transmit partially aggregated gradients, which could overflow. To achieve all-reduce compatibility, THC suggests a simple adaptation by increasing the number of bits $b\ge q$ for communication to accommodate the increasing value range. This leads to extra communication and is still not scalable for a larger number of workers.

\begin{figure*}[!htbp]
        \vspace{-0.1cm}
        \centering
        \begin{minipage}[t]{0.4\linewidth}{
		\vspace{-0.00in}
		\begin{center}
		\includegraphics[width=\textwidth, ]{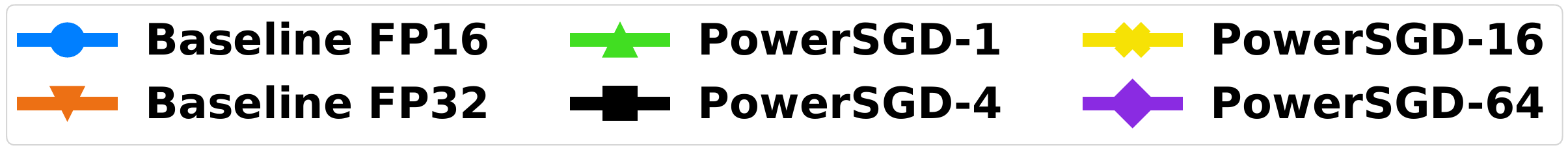}
		\end{center}
		}
        \end{minipage}

        \vspace{-2.3mm}
        \subfigure[BERT-large]{
		\begin{minipage}[t]{0.44\linewidth}{
		\vspace{-0.00in}
		\begin{center}
		\includegraphics[width=\textwidth, ]{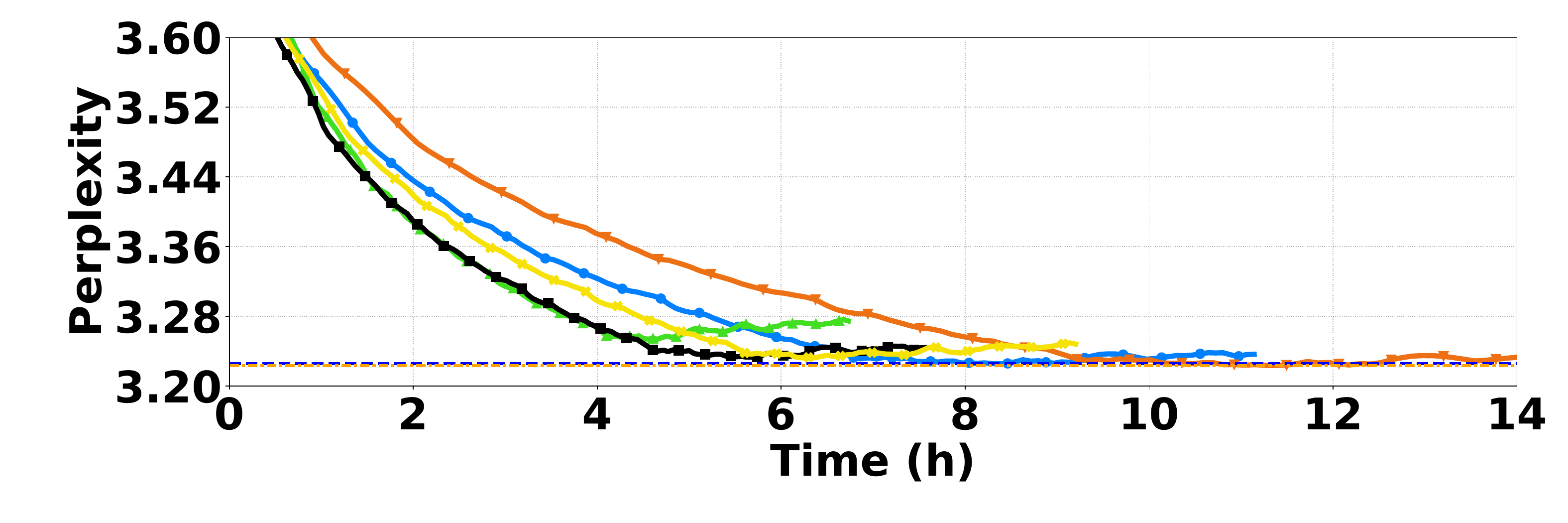}
		\end{center}
		}
            \vspace{-2.3mm}
		\label{subfig:topk-lm3}
		\end{minipage}
	}
	%
        %
        % \vspace{-2mm}
	\subfigure[VGG19]{
		\begin{minipage}[t]{0.44\linewidth}{
		\vspace{-0.00in}
		\begin{center}
		\includegraphics[width=\textwidth, ]{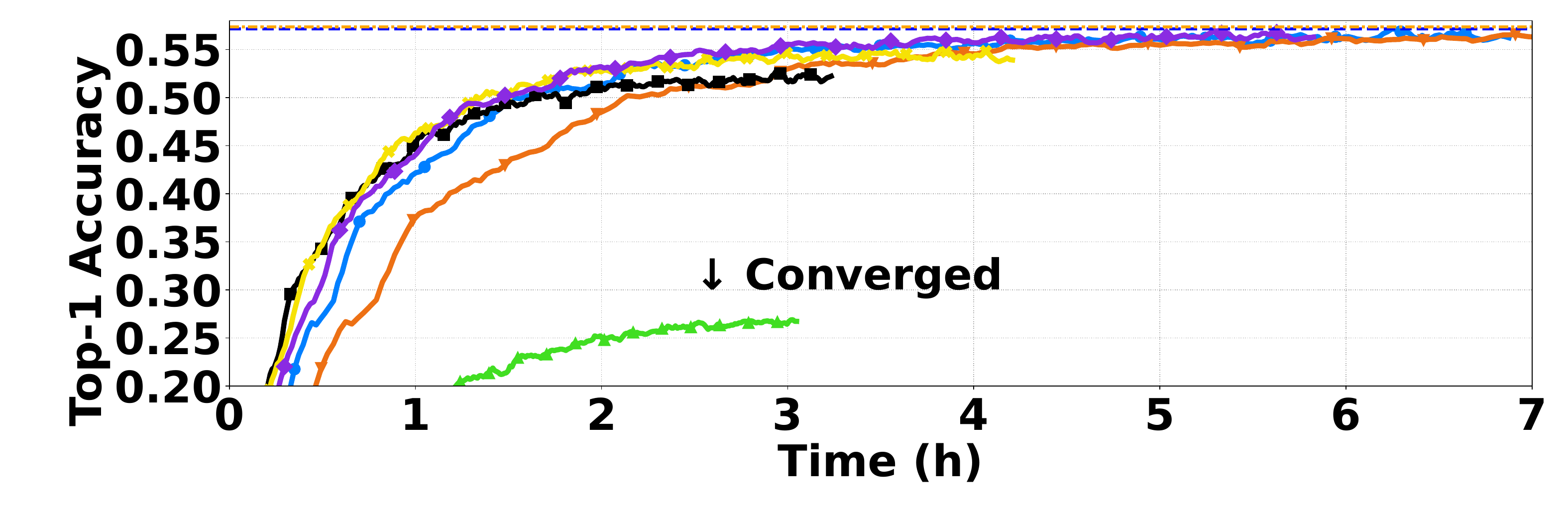}
		\end{center}
		}
            \vspace{-2.3mm}
		\label{subfig: topk-image3}
		\end{minipage}
	}
	\vspace{-0.5cm}
	\caption{The TTA of PowerSGD, altering the matrix rank $r$.}
        \vspace{-0.25cm}
	\label{fig: powersgd}
	
\end{figure*}

%\vspace{-0.1cm}
\subsubsection{Our proposed improvements}\label{subsubsec:thc-solutions} \quad
\vspace{0.1cm}

% \vspace{-0.05cm}
\subp{Partial rotation.}  
To alleviate the computational overhead of RHT, we observe that its recursive structure allows us to introduce an early stopping condition that takes into account the shared memory size of GPUs. Namely, for a vector of size $2^l$, the full RHT involves $l$ iterations. Instead, we stop the transform after $l' \le l$ iterations, picking the largest $l'$ such that the shared memory size is larger than $2^{l'}$. This is mathematically equivalent to dividing the gradient into $2^{l'}$-sized chunks and rotating each separately but with the advantage of using only a single GPU kernel to do so.

% limit th that performs a full randomized rotation of the gradient, we partition the gradient vector into chunks of size $2^{l'}$ and independently perform the chunk-based quantization and Hadamard Transformation. Chunk-based Hadamard Transformation is in essence a partial rotation $H_{l'}$, where $H_{l'}$ defined previously is the first $l'$ recursive steps of the full rotation $H_l$. 
% Here $l'$ is chosen such that $2^{l'} << d$ fits within the shared memory, where $d$ is the size of the gradient. 
% \MM{Above sentence needs clarification.}
After the partial rotation, we compute the value range $\max_p - \min_p$ of each chunk $p$, so that coordinates of higher magnitude only affect the quantization precision locally. 
Evaluation (Section~\ref{subsubsec:thc-evaluations}) shows that partial rotation achieves faster compression while remaining effective in reducing the value range to improve quantization precision.

%\smallskip
\subp{Saturation-based aggregation.} To achieve better all-reduce compatibility and handle overflows, we propose a {saturation-based} lossy aggregation that enables aggregation without extra communication overhead. {One insight is that RHT transforms the gradients $\nabla^{(i)}$ to a vector whose entries roughly follow a normal distribution $\mathcal N(0, 1 / ||\nabla^{(i)}||_2^2)$~\cite{vargaftik2021drive}. That is, the coordinates are strongly concentrated around $0$ and may cancel each other to some extent during the summation, reducing the chance of overflow. 
%and the remaining with large magnitudes are likely to cancel out with each other during aggregation, reducing the chance of overflow. 
}

We thus propose replacing summing gradients with the saturation operator $\text{Sat}(\cdot,\cdot)$ acting on top of the rotated and normalized gradients. Formally, 
%\vspace{-0.1cm}
%$
$ \text{Sat}(x, y) = \min(2^{b-1} - 1, \max(-2^{b-1}+1, x+y))$,
%$
where $x, y$ are $b$-bit for aggregation and $b=q$ in our experiments. We find that saturation does not introduce much error in our experiments because of the low probability of overflows after rotation and normalization. Other setups, e.g., using a large number of workers, may affect this conclusion. %We later show in end-to-end experiments~\ref{} that the benefit of the decreased communication overhead compared with the naive solution far outweighs the accuracy degradation.

As the number of workers $n$ increases, our saturation-based aggregation has to allocate more communication bits $b > q$ to upper-bound the probability of overflows. However,  the number of bits for saturation is generally less than the simple solution, as it takes advantage of the cancellation that arises from positive and negative values.  
% However, we argue that saturation-based aggregation still achieves better scalability than the naive solution. Intuitively this is because of the aforementioned cancellation between positive and negative values, which allows to reduce the communication overhead from $b+\lceil\log_2(n)\rceil$ to $b+\lceil\frac{1}{2}\log_2(n)\rceil$\Wenchen{TO check}. % and the normalized gradient values that concentrate around $0$.

Our techniques may generalize to other quantization sche\-mes, e.g., addressing integer summation overflow through saturation for~\cite{alistarh2017qsgd, bernstein2018signsgd, terngrad} and enhancing speed by replacing full RHT with partial rotation, e.g., for~\cite{suresh2017distributed, vargaftik2021drive}.

%https://ucl.zoom.us/j/2850478756

% \smallskip
%\vspace{0.02cm}
% \subp{Discussion.} We note that both techniques that we propose can potentially be generalized to other quantization-based compression works~\cite{suresh2017distributed, vargaftik2021drive, alistarh2017qsgd, bernstein2018signsgd, terngrad}. On the one hand, for any quantization-based compression, the aggregation can be abstracted as the integer summation given limited bits and hence faces the overflow issue. This can be addressed by our proposed saturation. Second, many works~\cite{suresh2017distributed, vargaftik2021drive} adopt RHT to optimize their quantization accuracy, and thus the partial rotation could potentially be applied to them as well.

%\noindent\textit{Potential future works.} The problem of handling overflows could also be abstracted as an estimator problem. 
%In an all-reduce topology, the estimator should fetch sources of inputs in predecessor workers and yield a $b$-bit representation that can later decoded as the estimation of the summed value, which may be transmitted to the successor workers as inputs. Potential estimators proposed by earlier works include Additive Error Counters~\cite{aee}, Multiplicative Error Estimator~\cite{mee}, etc., and we leave the adoption of these estimators as future works.

\vspace{-0.cm}
\subsubsection{Preliminary evaluation}\label{subsubsec:thc-evaluations}

We now present evaluations of our proposed optimization of THC, setting $b=q=4$. We compare against baseline THC that deploys full rotation of RHT and avoids overflows by using $b=8$ bits per coordinate.

\subp{Analysis of our design changes.} Table~\ref{tab:thc-throughput} shows that our proposed optimization effectively improves the throughput. First, the adoption of partial rotation contributes an up to $5.5\%$ increase in throughput ($b=4$), showing the successful reduction of the computational overhead. Second, the saturation-based aggregation achieves $50\%$ less communication, bringing up to $29.6\%$ higher throughput.

%\MM{It would have been better to separate these -- have each optimizaiton individually as well as together....}
Accordingly, the benefits of higher throughput are reflected in the improved end-to-end TTA results, as depicted in Figure~\ref{fig: thc}. 
In particular, by adding saturation and partial rotation to the original THC, TTA converges progressively faster to reach a given perplexity or top-1 accuracy.  
This, combined with the indistinguishable final perplexity/accuracy difference with the baselines, supports that both saturation and partial rotation incur little accuracy degradation, so the increase in throughput also leads to a better TTA.

Finally, we reduce the communication overhead by setting $b=q=2$. According to Table~\ref{tab:thc-throughput} and Figure~\ref{fig: thc}, the throughput is improved from $b=4$, but the TTA on BERT significantly degrades even compared to the Baseline FP16. This provides another piece of evidence that solely measuring the throughput is not adequate in evaluating end-to-end performance.

\vspace{-0.3cm}
\subsection{PowerSGD Low-rank Decomposition}\label{subsec:powersgd}

% \begin{figure}[t]
%         \centering
%         \begin{minipage}[t]{0.86\linewidth}{
% 		\vspace{-0.00in}
% 		\begin{center}
% 		\includegraphics[width=\textwidth, ]{figs/Powersgd_legend.pdf}
% 		\end{center}
% 		}
%         \end{minipage}

% 	\centering
% 	% \vspace{-0.135105in}
	
% 		\begin{minipage}[t]{0.44\linewidth}{
% 		\vspace{-0.00in}
% 		\begin{center}
% 		\includegraphics[width=\textwidth, ]{figs/lm_accuracy_4bit_or_16_testing_norotation2_tta.pdf}
% 		\end{center}
% 		}
% 		\label{subfig:topk-lm}
% 		\end{minipage}
% 	%
% 	%
% 	%
% 		\begin{minipage}[t]{0.44\linewidth}{
% 		\vspace{-0.00in}
% 		\begin{center}
% 		\includegraphics[width=\textwidth, ]{figs/accuracy_powersgd_16_test_acc_tta_imagenet.pdf}
% 		\end{center}
% 		}
% 		\label{subfig: topk-image}
% 		\end{minipage}
	
%         \vspace{-6mm}
%         \subfigure[BERT-large]{
% 		\begin{minipage}[t]{0.44\linewidth}{
% 		\vspace{-0.00in}
% 		\begin{center}
% 		\includegraphics[width=\textwidth, ]{figs/lm_accuracy_4bit_or_16_testing_norotation2_ttapart.pdf}
% 		\end{center}
% 		}
%             \vspace{-3mm}
% 		\label{subfig:topk-lm}
% 		\end{minipage}
% 	}
% 	%
%         \hspace{-2mm}
% 	\subfigure[VGG19]{
% 		\begin{minipage}[t]{0.44\linewidth}{
% 		\vspace{-0.00in}
% 		\begin{center}
% 		\includegraphics[width=\textwidth, ]{figs/accuracy_powersgd_16_test_acc_ttapart_imagenet.pdf}
% 		\end{center}
% 		}
%             \vspace{-3mm}
% 		\label{subfig: topk-image}
% 		\end{minipage}
% 	}
% 	\vspace{-0.1578975in}
% 	\caption{The TTA of PowerSGD, altering the matrix rank $r$.}
%     %\vspace{-0.1in}
% 	\label{fig: powersgd}
	
% \end{figure}
\begin{table}[t]
    \centering
    \resizebox{0.95\linewidth}{!}{
    \begin{tabular}{|c||c|c|c|c|c|c|c|c|c|} \hline
        \multirow{2}{*}{Task} & \multicolumn{2}{c|}{$r=1$} & \multicolumn{2}{c|}{$r=4$} & \multicolumn{2}{c|}{$r=16$} & \multicolumn{2}{c|}{$r=64$} \\ \cline{2-9}
        & $b$ & Thr. & $b$ & Thr. & $b$ & Thr. & $b$ & Thr. \\ \hline
        BERT & $0.0797$ & $5.49$ & $0.217$ & $4.89$ & $0.764$ & $4.01$ & $2.95$ & $3.03$ \\ \hline
        VGG19 & $0.0242$ & $21.0$ & $0.0872$ & $19.8$ & $0.339$ & $15.2$ & $1.36$ & $11.0$ \\ \hline
    \end{tabular}
    }
    \vspace{0.1cm}
    \caption{Bits-per-coordinate and throughput (in rounds per second) for PowerSGD, \mbox{varying the rank $r$.}}
    \vspace{-.641cm}
    \label{tab:powersgd-throughput}
\end{table}

We now analyze PowerSGD, a compression scheme based on low-rank decomposition. It is parameterized with a small integer $r$ indicating the target rank of approximated matrices. 
The results are shown in Figure~\ref{fig: powersgd} and Table~\ref{tab:powersgd-throughput}. Since PowerSGD is compatible with all-reduce~\cite{on-the-utility}, we instead discuss the issues of computational overhead and evaluation.

\smallskip
\subp{Computational overhead.} In Table~\ref{tab:powersgd-throughput}, we find that PowerSGD could achieve high com-
pression ratios, with up to $47\times$ (for $r=16$) less bits-per-coordinate than FP16. However, such a considerable communication reduction does not improve as much throughput as other compression schemes (Table~\ref{tab:topk-throughput} and ~\ref{tab:thc-throughput}). Moreover, increasing $r$ from $1$ to $64$ slows down the throughput by nearly half, with the communication overhead still being negligible. 
Profiling reveals that the major bottleneck is the \textit{overwhelmingly expensive} operation of matrix orthogonalization~\cite{qr-decomposition}, which consumes \textbf{\(39.7\%\)} and $47.4\%$ of the training time for BERT and VGG19 with $r=64$.

%\smallskip
\subp{The choice of metrics and baselines.}
{The results as depicted in Figure~\ref{fig: powersgd} provide another example of how $r=1$ achieves a higher throughput than $r=16$ and $64$ but converges slower and to a lower accuracy for VGG19, highlighting the importance of TTA evaluation. Also, PowerSGD $r=4$ significantly outperforms Baseline FP32 but only provides a modest benefit compared to Baseline FP16, highlighting the importance of using a stronger baseline.}
% \Wenchen{Note: Throughput by commenting out the orthogonalization is: $5.026$, $20.9130$ respectively.}

%\subp{} To conclude, decreasing the rank \( r \) improves throughput but degrades TTA and final accuracy; also, a correct choice of \( r \) provides only a modest benefit compared to Baseline FP16.

%\smallskip
%\subp{Discussion .} %The experimental results of PowerSGD provide another piece of evidence that the current common practices of evaluation are insufficient. Regarding the evaluation metric, 
%In this case study, decreasing the matrix rank $r$ improves the throughput. However, once again, the higher throughput does not necessarily translate to better TTA. For example, decreasing $r$ to $1$ on VGG19 results in much lower TTA and lower final accuracy, \ie, it takes longer to reach and stop at lower accuracy. Regarding the choice of the baseline, we observe that PowerSGD generally outperforms Baseline FP32 by a significant margin, but the margin diminishes when compared with Baseline FP16. 

\vspace{-0.1cm}
\section{Conclusions and Discussion }\label{sec:conclusions}
\vspace{-0.01cm}
This paper provides insights into the issues involved in designing and evaluating state-of-the-art gradient compression systems. By considering the challenges in measuring performance, going beyond simple but insufficient measures such as throughput to the more robust measure of time to accuracy, we aim to shed light on designs that enhance the utility of gradient compression. Specifically, our case study describes how we identify problems and solve them with this framework in mind.

We note, however, that perhaps time to accuracy is itself not the only appropriate metric.
Key considerations not taken into account with TTA are the overall power used or the overall cost to construct the model. We leave the issue of determining a framework that takes power and/or cost into account as an exciting future direction.  We believe the approach suggested here, of focusing on the right metrics to judge end-to-end performance, would be very important to comparing compression schemes in these contexts.  

\paragraph{Acknowledgments:} We thank the anonymous reviewers for their insightful comments and suggestions. Michael Mitzenmacher was supported in part by NSF grants CCF-2101140,
CNS-2107078, and DMS-2023528.

% \subp{Building an aggregation estimator for quantization-based compression.} In quantization-based compression schemes like THC, the ways to handle overflow (such as via saturation) can be abstracted as the problem that estimates the summation of arrays of bounded integers in $[0, 2^{b}-1]$. In an all-reduce topology, such an estimator, as denoted as $\oplus$, should fetch sources of inputs for summation and yield a $b$-bit representation that can later decoded as the estimation of the summed value. The output representation may also be transmitted to the downstream hops as inputs of the estimator. Potential estimators that are proposed by earlier works include Additive Error Counters~\cite{}, Multiplicative Error Estimator~\cite{}, etc. 

%\newpage

\bibliographystyle{ACM-Reference-Format} 
\bibliography{hotnets24CR}

\end{document}